\documentclass{article}

\usepackage{arxiv}

\usepackage[utf8]{inputenc} 
\usepackage[T1]{fontenc}    
\usepackage{hyperref}       
\usepackage{url}            
\usepackage{booktabs}       
\usepackage{amsfonts}       
\usepackage{nicefrac}       
\usepackage{microtype}      
\usepackage{lipsum}
\usepackage{graphicx}

\title{Successive Subspace Learning: An Overview}

\author{%
  Mozhdeh Rouhsedaghat \\
  Department of Electrical Engineering\\
  University of Southern California\\
  Los Angeles, California, USA\\
  \texttt{rouhseda@usc.edu} \\
  \And
  Masoud Monajatipoor \\
  Department of Electrical Engineering\\
  University of California, Los Angeles\\
  Los Angeles, California, USA\\
  \texttt{monajati@ucla.edu} \\
  \AND
  Zohreh Azizi \\
  Department of Electrical Engineering\\
  University of Southern California\\
  Los Angeles, California, USA\\
  \texttt{zazizi@usc.edu} \\
  \And
  C.-C. Jay Kuo \\
  Department of Electrical Engineering\\
  University of Southern California\\
  Los Angeles, California, USA\\
  \texttt{jckuo@usc.edu} \\
 
}

\begin{document}

\maketitle

\begin{abstract}

Successive Subspace Learning (SSL) offers a light-weight unsupervised feature learning method based on inherent statistical properties of data units (e.g. image pixels and points in point cloud sets). It has shown promising results, especially on small datasets. In this paper, we intuitively explain this method, provide an overview of its development, and point out some open questions and challenges for future research. 

\end{abstract}

\section{Introduction}\label{sec:introduction}

In recent years, deep convolutional neural networks have achieved superior performance in many fields and are considered the state-of-the-art method for feature learning from images and videos. However, they are known to be data-hungry, meaning that they require a large amount of annotated training data to learn effective features, and may not be as effective when are trained on small datasets. Furthermore, training deep neural networks (DNNs) is computationally expensive and time-consuming, thus imposing major challenges in edge and mobile computing. 

SSL offers a light-weight unsupervised feature learning method based on inherent statistical properties of data units (e.g. image pixels and points in point cloud sets). It has been applied to different types of data, e.g. images~\cite{chen2020pixelhop++}, point-clouds~\cite{zhang2020pointhop++}, voxels~\cite{liu2021voxelhop}, etc. The number of SSL model parameters is significantly lower than that of DNNs~\cite{rouhsedaghat2020facehop, rouhsedaghat2020low}, and its parameters can be computed without using back-propagation, so it is computationally efficient and can be trained on CPU. SSL can extract powerful features from training data as evidenced by its superior performance compared with DNNs when the number of training data is small~\cite{zhang2020pointhop++, rouhsedaghat2020facehop,liu2021voxelhop}. 

The rest of the paper is organized as follows. An intuitive explanation about SSL is presented in Sec. \ref{sec:approach}. Existing SSL-based models are reviewed in Sec.~\ref{sec:app}. Several potential future research directions are discussed in Section~\ref{sec:research}. 

\section{Approach}\label{sec:approach}

The main idea of SSL is computing the weights of its model by a closed-form expression without using back-propagation. Kuo {\em et al.} developed the idea of SSL in \cite{kuo2016understanding,kuo2017cnn,kuo2018data} and introduced the Saab transform \cite{kuo2019interpretable}. Recently, Chen {\em et al.} \cite{chen2020pixelhop++} developed this idea further, and introduced PixelHop++ which is considered as the latest SSL-based method for feature learning from images. In this section, we focus on PixelHop++ because the proposed SSL-based methods for feature learning from other types of data have a similar foundation. 

Inspired by the architecture of DNNs, PixelHop++ is a multi-level model, in which each level consists of \textbf{PixelHop++ unit} followed by a max-pooling layer. 

\begin{figure}
  \centering
  \includegraphics[scale=.34]{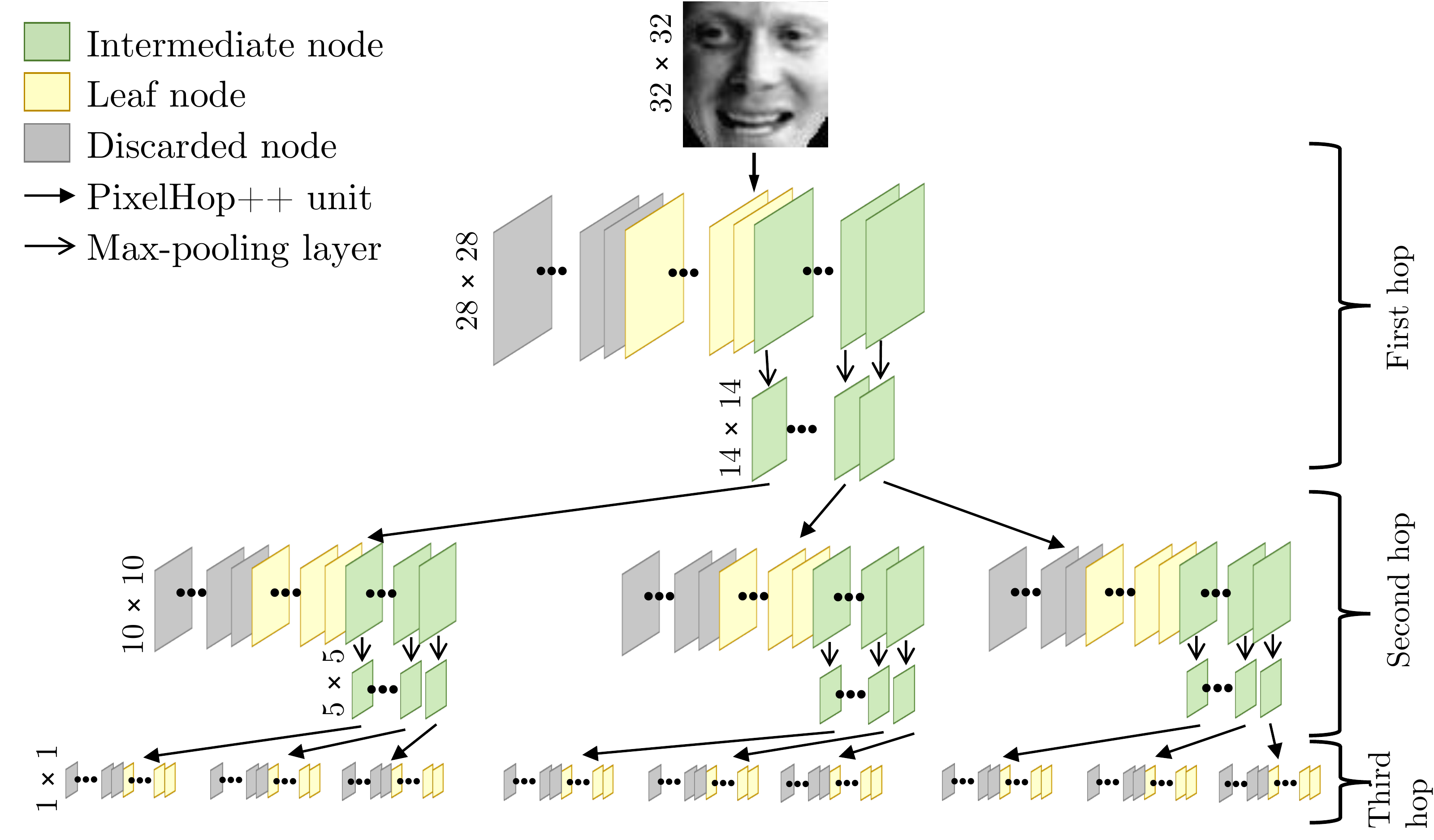}
\caption{: Illustration of data flow in the three-level PixelHop++
model. Image taken from \cite{rouhsedaghat2020low} with permission from
the authors.}
\end{figure}

In the training phase of a PixelHop++ unit, the goal is to provide effective projection vectors (kernels) to extract strong features from image patches. In the first step, a window of a certain size slides through the image, and local patches are extracted. Then, the projection kernels are computed as follows:

\begin{itemize}
\item The first projection kernel is the mean filter (i.e., \begin{math}\frac{1}{\sqrt{n}}\times (1, 1, ..., 1) \end{math}). By projecting all patches of an image on this vector, their mean values are computed and stored in one output channel (node) called the DC channel. 
\item For patches of size $m \times m \times 1$, after extracting the mean, we can still define $m^2$-1 independent projection kernels for feature learning from residuals. In PixelHop++, Principal Component
Analysis (PCA) is used to compute the principal components of the whole training patches. Computed principal components are used as the projection kernels and by projecting all patches of an image on each of these principal components, one output channel called an AC channel is formed. 
\end{itemize}

After computing the kernels and projecting image patches on them, we will have 1 DC channel and $m^2$-1 AC channels.  The intuition behind using PCA is that the first k principal components are the k projection kernels that data varies most in their direction and can explain the data best. 

Then, there is one step of data pruning to reduce the number of output channels and remove unnecessary information. We call the variance of training input patches along each of the kernels the "energy'' of the kernel (or the energy of the corresponding channel) and use its ratio as a metric for removing some of the output channels. Intuitively, if the variation of the data along one axis is too small, it means the axis does not carry much information and can be removed. Two energy ratio thresholds are defined in the PixelHop++ architecture called $E_f$ and $E_c$ ($E_c$<=$E_f$). If the energy ratio of a channel is less than $E_c$, it will be discarded (discarded nodes/channels) and if it is more than $E_f$ it is forwarded to the next level for further energy compaction (intermediate nodes/channels). Nodes with an energy ratio between $E_c$ and $E_f$ are kept in the current level (Leaf nodes/channels).

Each output intermediate channel of a PixelHop++ unit serves as the input of a separate PixelHop++ unit in the next level because they are not correlated with other channels.  Note that having a multi-level architecture increases the receptive field of each computed response at the higher levels and is important for improving the image classification accuracy.  

Finally, all output channels in each level can be used for generating the final feature vector. Channels at lower levels provide a more detailed representation of the input image while channels at higher levels yield a more abstract representation (with a larger receptive field) but contain less detail. 

\section{Other SSL-based Models and Applications}\label{sec:app}

FaceHop~\cite{rouhsedaghat2020facehop} uses PixelHop++ for low-resolution face gender classification. For generating the final feature vector from the PixelHop++ output channels, it defines several regional responses corresponding to facial salient regions and applies PCA to them to remove the spatial correlation of features inside each channel. Then, for each region, it concatenates the dimension-reduced responses of different channels to train a separate classifier. Finally,
a meta classifier is used to ensemble the results of individual classifiers. FaceHop achieves a better result compared with LeNet-5 when both are trained on small-size datasets while its model size is smaller. 

Rouhsedaghat {\em et al.}~\cite{rouhsedaghat2020low} used PixelHop++ for low-resolution face recognition. While having a model of $79\times$ smaller size, they achieved results close to the state-of-the-art. Their proposed model contains a block for generating pairwise features from PixelHop++ output channels for each pair of face images. The mentioned block uses the cosine similarity and the L2-norm ratio of several regional responses (corresponding to facial salient regions) in each channel to generate the final feature vector. 

PointHop++ \cite{zhang2020pointhop++} is the latest SSL-based method for feature learning from 3D point cloud. It achieves state-of-the-art classification results while having a smaller model size compared with previous models. The basis of PointHop++ is similar to PixelHop++, but instead of using a sliding window for extracting initial patches, it uses k-Nearest-Neighbours (kNN) to retrieve neighbor points. After training the PointHop++ model, the linear least squares regression (LLSR) is trained on the obtained features to calculate the final probability vector. 

Lei {\em et al.} \cite{lei2020nites} proposed an SSL-based texture synthesis method called NITES. It is mathematically transparent and efficient in training and inference. The input is a single exemplary texture image. NITES crops out patches from the input and analyzes the statistical properties of these texture patches to obtain their joint spatial-spectral representations. Then, the probabilistic distributions of samples in the joint spatial-spectral spaces are characterized. Finally, texture images that are visually similar to the exemplary texture image can be generated automatically. 

VoxelHop \cite{liu2021voxelhop} is an SSL-based method for feature learning from 3D imaging data. It outperforms state-of-the-art 3D CNN-based classification methods on 3D medical imaging data. VoxelHop uses a sliding window of size $m\times m\times C_0$ for extracting patches from the input to each PixelHop++ unit. After training the PixelHop++ units, it uses a Label Assisted Regression (LAG) unit~\cite{chen2020pixelhop} to apply supervised dimension reduction on the output of max-pooling layers. Finally, it concatenates the dimension-reduced responses to form the final feature vector. 

\section{Future Research and Open Questions}\label{sec:research}

SSL provides an effective unsupervised data embedding method where the data can be images, videos, and 3D point clouds, etc. It avoids the end-to-end optimization as formulated in DNNs and decomposes a machine learning model into two separate modules: ``unsupervised feature learning" and ``supervised decision learning". Generally, the modularized approach is more scalable and easier to debug than the end-to-end approach. 

It is worthwhile to revisit many image processing and computer vision problems, which are well solved by the deep learning approach, using the emerging SSL approach for lower computational complexity and memory requirements. The list includes image classification, super-resolution, denoising, semantic segmentation, generation, retrieval, compression, object detection and tracking, etc. Although DNNs have achieved great success in image-based processing, it is still challenging for deep learning to be applied to general video processing due to the huge amount of 3D spatial-temporal data. It appears that the SSL approach can be an attractive alternative. 

Besides the above-mentioned applications, it is important and interesting to investigate some fundamental problems associated with SSL. For example, how to integrate SSL-based feature learning and traditional machine learning for incremental learning, transfer learning, adversarial learning, weakly-supervised learning, and few-shot (or zero-shot) learning.

\bibliographystyle{unsrt}  
\bibliography{file}  

\end{document}